\pdfoutput=1

\documentclass[11pt]{article}

\usepackage[]{acl}

\usepackage{multicol}
\usepackage{multirow}
\usepackage{array}
\usepackage{booktabs}

\usepackage{times}
\usepackage{latexsym}

\usepackage[T1]{fontenc}

\usepackage[utf8]{inputenc}

\usepackage{microtype}

\usepackage{inconsolata}

\usepackage{graphicx}

\title{\raisebox{-1.5ex}{\includegraphics[width=1cm]{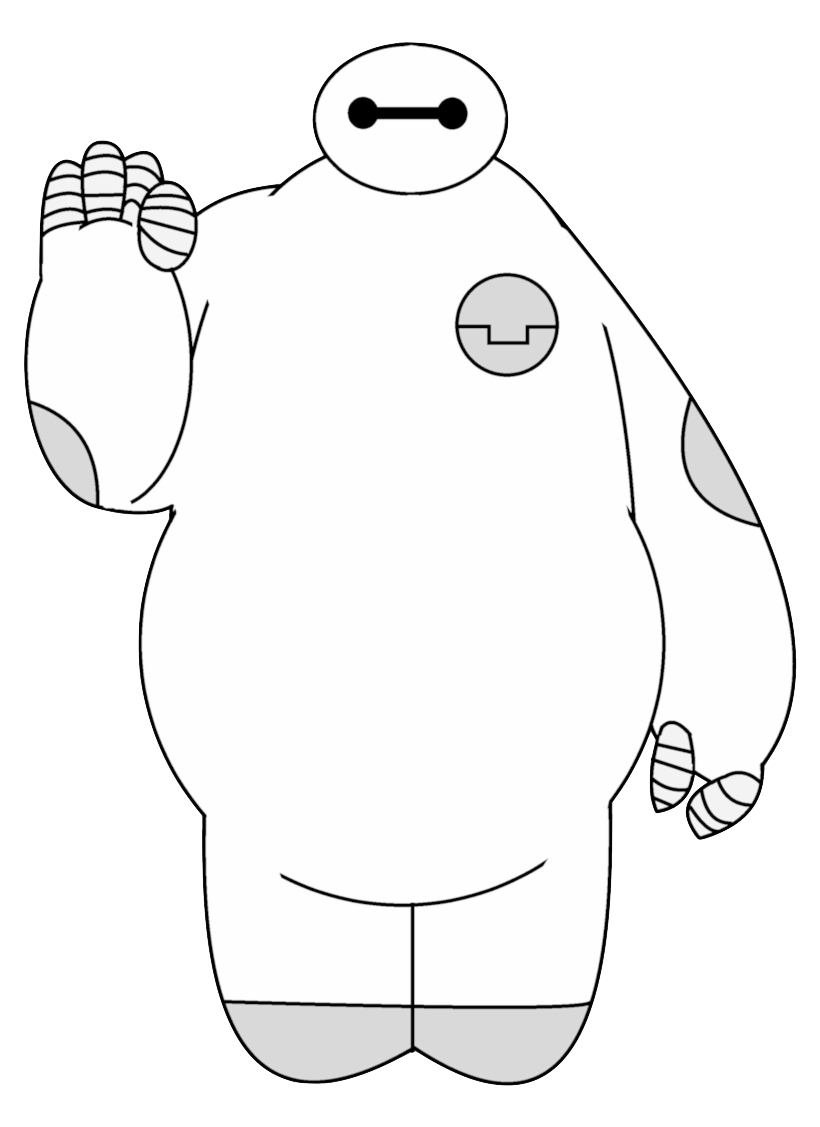}}A Survey of LLM-based Agents in Medicine:  \\  How far are we from Baymax?}

\author{Wenxuan Wang$^{1}$\thanks{~~Wenxuan Wang and Zizhan Ma are equal contribute to this paper.} \quad Zizhan Ma$^{1}$$^*$ \quad Zheng Wang$^1$ \quad Chenghan Wu $^{1}$ \\ \bf \bf Jiaming Ji$^{2}$ \quad \bf Wenting Chen$^{3}$  \quad  \bf Xiang Li$^4$  \quad \bf Yixuan Yuan$^1$ \\
$^1$The Chinese University of Hong Kong  \\
$^2$ Peking University  \quad \quad  $^3$ City University of Hong Kong\\
$^4$ Massachusetts General Hospital and Harvard Medical School\\
$^1$\texttt{\{wenxuanwang,zizhan.ma\}@link.cuhk.edu.hk}   \quad
$^2$\texttt{wentichen7-c@my.cityu.edu.hk} \\ 
}

\begin{document}
\maketitle

\begin{abstract}
Large Language Models (LLMs) are transforming healthcare through the development of LLM-based agents that can understand, reason about, and assist with medical tasks. This survey provides a comprehensive review of LLM-based agents in medicine, examining their architectures, applications, and challenges. We analyze the key components of medical agent systems, including system profiles, clinical planning mechanisms, medical reasoning frameworks, and external capacity enhancement. The survey covers major application scenarios such as clinical decision support, medical documentation, training simulations, and healthcare service optimization. We discuss evaluation frameworks and metrics used to assess these agents' performance in healthcare settings. While LLM-based agents show promise in enhancing healthcare delivery, several challenges remain, including hallucination management, multimodal integration, implementation barriers, and ethical considerations. The survey concludes by highlighting future research directions, including advances in medical reasoning inspired by recent developments in LLM architectures, integration with physical systems, and improvements in training simulations. This work provides researchers and practitioners with a structured overview of the current state and future prospects of LLM-based agents in medicine.

\end{abstract}

\section{Introduction}

Large Language Models (LLMs) are changing the field of artificial intelligence with their strong capabilities in text understanding, generation, and reasoning. The development of LLM-based agents has achieved notable success in many areas, from creative writing \cite{10.1145/3490099.3511105} to complex decision-making \cite{chai2025empowering,wei2023chainofthoughtpromptingelicitsreasoning}, which opens new opportunities for automating and enhancing human expertise. These agents have been applied in various fields by using the ability of LLMs to process and analyze complex information \cite{Wang_2024,cheng2024exploringlargelanguagemodel,xi2023risepotentiallargelanguage}.

In the medical domain, LLM-based agents have improved several clinical tasks. Recent work shows their use in diagnostic support \cite{kim2024mdagentsadaptivecollaborationllms}, patient communication \cite{mukherjee2024polarissafetyfocusedllmconstellation}, and medical education \cite{yu2024aipatientsimulatingpatientsehrs}. By combining LLMs with medical knowledge bases, clinical guidelines, and healthcare systems, these agents are designed to understand complex medical situations \cite{wei2024medaideomnimedicalaide}, offer evidence-based recommendations \cite{tang2024medagentslargelanguagemodels}, and support healthcare delivery \cite{mukherjee2024polarissafetyfocusedllmconstellation}. Despite these advances, the field still faces several challenges, including implementation issues \cite{sun2024applications}, safety concerns \cite{yuan-etal-2024-r}, and ethical considerations \cite{yan2025application}. Addressing these challenges is essential for the safe and reliable integration of LLM-based agents into clinical practice. Therefore, a comprehensive review is needed to analyze the current status and future directions of LLM-based agents in medicine.

In this article, we provide a systematic review of LLM-based agents in medicine, examining important research questions and future directions. We first discuss the architectures and methods, including system profile, external capacity enhancement, clinical planning, and medical reasoning in Section~\ref{sec:architectures}. Section~\ref{sec:applications} covers the various clinical and administrative application scenarios in which these agents are used. Section~\ref{sec:evaluation} outlines the evaluation frameworks and metrics for assessing their performance in healthcare settings. Finally, Section~\ref{sec:discussion} highlights key challenges and future research directions for improving the reliability, safety, and clinical integration of LLM-based agents. 

This review analyzed 60 studies on LLM-based medical agents published between 2022-2024, selected from major databases using healthcare AI-related keywords. The initial search yielded 300 papers, narrowed to 80 after screening, with 60 meeting final inclusion criteria.


\section{Background} \label{sec:background}

This section outlines the core differences between \textbf{LLMs} and \textbf{LLM-based agents} and highlights the unique considerations required for deploying such agents in medicine.

\subsection{LLM vs. LLM-based Agent}

An agent, as defined in AI, perceives its environment and takes actions accordingly \cite{russell2016artificial}. An LLM-based agent extends traditional LLMs by integrating external knowledge retrieval, task planning, and tool invocation, enabling structured decision-making in real-world applications \cite{xi2023risepotentiallargelanguage}. Unlike standard LLMs, which primarily process text, these agents operate autonomously and adapt dynamically to new information and tasks.

\subsection{Unique Considerations for LLM-based Agents in Medicine}

Deploying LLM-based agents in healthcare requires addressing several critical factors:

\noindent \textbf{Multimodal Integration}. Medical data spans text, imaging, and laboratory results. Agents must process and synthesize these inputs for accurate decision support \cite{electronics13112002}.

\noindent \textbf{Clinical Collaboration}. Healthcare relies on interdisciplinary work. Agents should facilitate information sharing and human-AI collaboration, ensuring physicians maintain oversight \cite{strong2024humanaicollaborationhealthcareguided}.

\noindent \textbf{Accuracy and Reliability}. Given the impact on patient outcomes, these agents must meet strict validation standards and minimize errors in diagnosis and treatment \cite{reddy2024generative}.

\noindent \textbf{Transparency and Traceability}. Clinical decisions must be auditable and explainable to align with medical ethics and regulatory requirements \cite{10.3389/frai.2022.879603}.

\begin{figure*}[h]
    \centering
    \includegraphics[width=\textwidth]{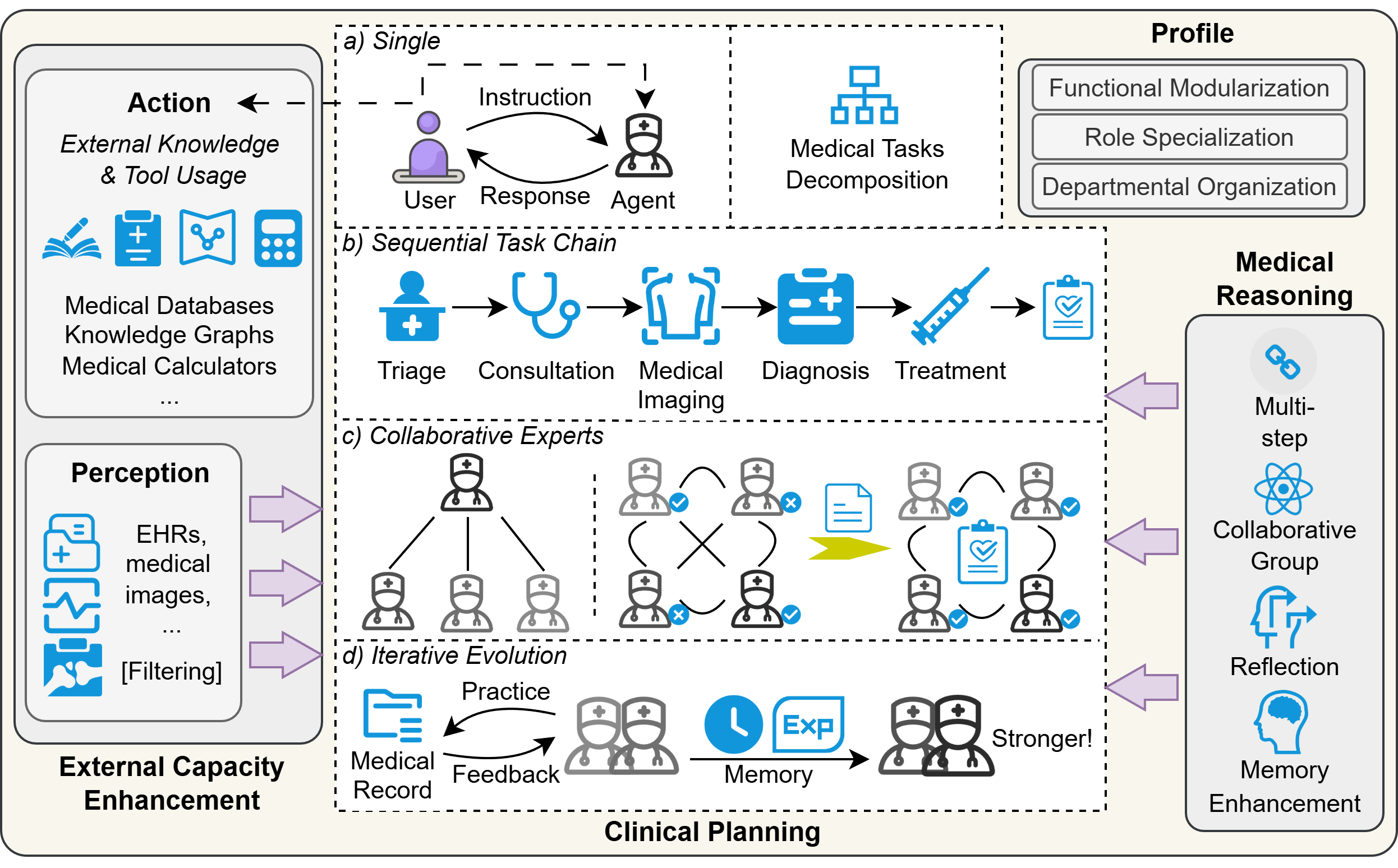} 
    \caption{Conceptual framework of LLM-based medical agents. This figure depicts the architecture of the proposed LLM-based Medical Agent, consisting of system profile, external capacity enhancement, clinical palnning and medical reasoning. It supports four agent paradigms: \textbf{a) Single Agent}, \textbf{b) Sequential Task Chain}, \textbf{c) Collaborative Experts}, and \textbf{d) Iterative Evolution}. The framework integrates external tools and reasoning mechanisms to enable applications in medicine.
}
    \label{fig:framework_diagram}
\end{figure*}

\section{LLM-based Medical Agent Architecture} \label{sec:architectures}

LLM-based agents in medicine require well-defined architectures to integrate complex clinical knowledge, facilitate medical decision-making, and ensure safe and effective deployment. This section presents a systematic overview of their architectural components, focusing on how these agents structure their operations to enhance clinical performance.

\subsection{Profile}

The profile of an agent plays a key role in defining and managing its role attributes, behavioral patterns, and operational competencies within medical systems, which traditionally involve information dissemination, resource distribution, and quality assurance. In medical applications, agent profiles follow three prototypes:

\noindent \textbf{Functional Modularization.}
This approach structures the agent system into specialized functional modules, each responsible for distinct tasks such as clinical data analysis or diagnostic reasoning. Systems like \textit{MEGDA} \cite{baniharouni2024magdamultiagentguidelinedrivendiagnostic} implement function-driven profiles where task assignments and workflows are explicitly defined to improve efficiency and adaptability.

\noindent \textbf{Role Specialization}
By mirroring real-world medical roles, this paradigm assigns agents to specific clinical functions, including diagnosis, medical imaging, treatment planning, and surgical assistance. These agents incorporate domain-specific knowledge and interact with healthcare systems for tasks such as imaging analysis and interdisciplinary coordination. In agent-driven operating room simulations \cite{wu2024surgboxagentdrivenoperatingroom}, LLM-based agents take on distinct medical roles to support clinical decision-making.

\noindent \textbf{Departmental Organization}
This framework structures agents based on medical disciplines, such as cardiology or hematology, establishing domain-specific knowledge boundaries. Agents rely on specialized disease knowledge graphs and dynamic collaboration mechanisms to facilitate interdisciplinary consultations. In multi-agent medical applications \cite{tang2024medagentslargelanguagemodels}, profiles are defined to reflect departmental expertise, improving coordination and decision-making in complex medical scenarios.

\subsection{Clinical Planning}

Effective clinical planning is at the core of LLM-based medical agents. The planning process breaks down complex medical tasks into smaller subtasks so that the system can interact with tools and databases specific to each clinical area \cite{Mehandru2024EvaluatingLL}. This division of tasks improves operational efficiency and aids in locating and correcting errors.

\noindent \textbf{Task Decomposition}
Clinical planning often follows a structured decomposition from high-level objectives to specific actions. A \textit{Single-Agent} model handles tasks autonomously, while a \textit{Sequential Task Chain} approach structures planning into distinct steps, such as data ingestion, hypothesis generation, treatment planning, and risk assessment. Each step interacts with specialized medical tools, ensuring task separation and facilitating precise error correction \cite{liu2024medchainbridginggapllm}.

\noindent \textbf{Multi-Agent Collaboration Across Departments}
For complex cases requiring interdisciplinary expertise, a \textit{Collaborative Experts} model assigns specialized agents to clinical areas such as radiology, pathology, and laboratory analysis. These agents communicate using standardized protocols to aggregate findings and refine diagnoses. This reduces diagnostic uncertainty by integrating insights from multiple specialties \cite{tang2024medagentslargelanguagemodels}.

\noindent \textbf{Adaptive Planning Architecture}
A dynamic \textit{Sequential Task Chain} or \textit{Collaborative Experts} framework adjusts decision-making based on real-time data and task complexity. For example, MDAgents framework \cite{kim2024mdagentsadaptivecollaborationllms} employs LLMs with predefined medical roles, which function autonomously or in coordination. Planning layer continuously updates clinical strategies, prioritizes urgent cases, and refines past decisions based on new evidence. Federated learning mechanisms further enhance adaptability by integrating diverse clinical experiences \cite{dutta2024adaptivereasoningactingmedical}.

\noindent \textbf{Iterative Self-Evolution}
Beyond static workflows, \textit{Iterative Evolution} frameworks enable continuous improvement. These systems maintain an experience base of past cases, refining decision-making over time. Self-improvement mechanisms allow agents to autonomously incorporate new medical data and learn from previous outcomes, progressively enhancing accuracy and reliability \cite{li2024agenthospitalsimulacrumhospital, du2024llmssimulatestandardizedpatients}.

\subsection{Medical Reasoning}

The Medical Reasoning module enhances diagnostic accuracy and transparency by structuring logical inference processes and integrating real-time feedback.

\noindent \textbf{Multi-Step Diagnostic Reasoning}
Complex cases are analyzed through sequential inference, where Chain-of-Thought methods \cite{wei2023chainofthoughtpromptingelicitsreasoning} generate step-by-step reasoning, and Tree-of-Thought approaches \cite{yao2023treethoughtsdeliberateproblem} explore multiple hypotheses in parallel, discarding less probable options. This structured approach improves diagnostic precision \cite{dutta2024adaptivereasoningactingmedical}.

\noindent \textbf{Reflective Decision-Making}
To handle clinical uncertainty, the system iteratively refines conclusions by incorporating real-time feedback and expert input. Inspired by the ReAct framework, it alternates between reasoning and action, identifying inconsistencies and improving decision robustness \cite{yao2023reactsynergizingreasoningacting,yue2024clinicalagentclinicaltrialmultiagent}.

\noindent \textbf{Collaborative Group Reasoning}
A multi-agent reasoning framework assigns specialized agents—such as primary care providers and specialists—to perform independent analyses. Their conclusions are aggregated through consensus mechanisms, mitigating biases and enhancing reliability \cite{zuo2025kg4diagnosishierarchicalmultiagentllm}.

\noindent \textbf{Memory-Enhanced Reasoning}
Integrating long-term memory modules enables agents to accumulate medical knowledge and past clinical experiences, refining decision-making over time. This persistent memory allows the system to adapt to new medical insights, improve reasoning capabilities, and maintain continuity in patient care \cite{li2024agenthospitalsimulacrumhospital}. Additionally, experience-based learning mechanisms enable LLM-based agents to update their diagnostic strategies dynamically, leading to more context-aware and personalized medical insights \cite{jiang2024longtermmemoryfoundation}.

\subsection{External Capacity Enhancement}

The external capacity enhancement augments the agent's capabilities by integrating it with real-world clinical data sources and specialized tools.

\noindent \textbf{Perception} subsystem processes diverse clinical inputs, including structured electronic health records (EHRs) to access patient histories and clinical parameters. Advanced Optical Character Recognition (OCR) techniques convert scanned documents into text, while models like CLIP analyze medical images, facilitating comprehensive multimodal understanding.

\noindent \textbf{Knowledge Integration} connects the agent with external sources such as medical knowledge graphs, drug interaction databases, and clinical guideline repositories. This connection helps the agent verify its inferences with trusted sources, thereby increasing both its accuracy and reliability \cite{li2024mmedagentlearningusemedical,huang2024benchmarkinglargelanguagemodels}.

\noindent \textbf{Action} layer allows agents to perform clinical tasks by using specialized tools such as medical calculators, electronic health record interfaces, and image analysis software. The system calls additional functionalities when processing complex data, ensuring that its outputs are complete and take the context into account \cite{shi2024ehragentcodeempowerslarge,zhu2024mentibridgingmedicalcalculator}.

\section{Application Scenarios} \label{sec:applications}

LLM-based agents are applied in various areas of medicine. This section outlines the main application scenarios and provides a summary in Table~\ref{tab:appliction}.\begin{table*}[h]
\centering
\caption{Summary of application of LLM-Agents in the medical field.}
\renewcommand\arraystretch{1.3}
\scalebox{0.65}{
\begin{tabular}{c|cc|c|c|c}
\toprule
    \textbf{Purpose} & \multicolumn{2}{|c|}{\textbf{Functionality}} & \textbf{Work} & \textbf{Framework Type} & \textbf{Tool Use}\\
    \cline{1-6}
    \multirow{10}{8em}{Clinical Decision Support and Diagnosis} & \multicolumn{2}{|c|}{Refine Diagnostic Reasoning} & \cite{dutta2024adaptivereasoningactingmedical} & Adaptive Planning & -\\
    \cline{2-6}
     & \multicolumn{2}{|c|}{Reduce Cognitive Bias} & \cite{info:doi/10.2196/59439} & Collaborative Experts & -\\
    \cline{2-6}
     & \multicolumn{2}{|c|}{Task Coordination} & \cite{wei2024medaideomnimedicalaide} & Sequential Task Chain & - \\
    \cline{2-6}
     & \multicolumn{2}{|c|}{\multirow{2}{*}{Diagnosis Accuracy}} & \cite{kim2024demonstrationadaptivecollaborationlarge} & Collaborative Experts & Yes\\ 
    \cline{4-6}
     & && \cite{tang2024medagentslargelanguagemodels} & Collaborative Experts  & - \\
    \cline{2-6}
     & \multirow{3}{8em}{Domain Specific Functionalities} & \multicolumn{1}{|c|}{Clinical Trial Outcome Prediction} &\cite{yue2024clinicalagentclinicaltrialmultiagent} & Sequential Task Chain & Yes \\ 
    \cline{3-6}
     & & \multicolumn{1}{|c|}{Patient Interaction Safety} & \cite{mukherjee2024polarissafetyfocusedllmconstellation} & Sequential Task Chain & - \\ 
    \cline{3-6}
     & & \multicolumn{1}{|c|}{Prescription Validation} & \cite{van2024rxstrategistprescriptionverification} & Sequential Task Chain & Yes \\ 
    \cline{2-6}
     & \multicolumn{2}{|c|}{Diagnosis Capability} & \cite{yan2024clinicallabaligningagentsmultidepartmental} & Collaborative Experts & - \\
    \cline{2-6}
     & \multicolumn{2}{|c|}{Integrated Modelling} &\cite{fan2024aihospitalbenchmarkinglarge} & -- & - \\ 
    \cline{1-6}
    \multirow{6}{8em}{Clinical Data Analysis and Documentation} & \multicolumn{2}{|c|}{Mortality Prediction} & \multirow{2}{*}{\cite{wang2024colacareenhancingelectronichealth}} & \multirow{2}{*}{Collaborative Experts} & \multirow{2}{*}{Yes}\\ 
    \cline{2-3}
     & \multicolumn{2}{|c|}{Hospital Readmission Analysis} & & & \\
    \cline{2-6}
     & \multicolumn{2}{|c|}{Clinical Documentation} & \cite{lee2024improvingclinicaldocumentationai} & Single Agent & - \\
    \cline{2-6}
     & \multicolumn{2}{|c|}{Patient Friendly Medical Reports} & \cite{sudarshan2024agenticllmworkflowsgenerating} & Iterative Evolution & Yes \\
    \cline{2-6}
     & \multicolumn{2}{|c|}{Integrated Simulation} & \cite{li2024agenthospitalsimulacrumhospital} & Iterative Evolution & - \\
    \cline{1-6}
    \multirow{6}{8em}{Medical Training and Simulation} & \multicolumn{2}{|c|}{Evaluated Diagnosis and Treatment Performance} & \cite{yan2024clinicallabaligningagentsmultidepartmental} & Collaborative Experts & - \\
    \cline{2-6}
     & \multicolumn{2}{|c|}{\multirow{2}{*}{Integrated Simulation}} & \cite{fan2024aihospitalbenchmarkinglarge} & -- & - \\
    \cline{4-6}
     & && \cite{li2024agenthospitalsimulacrumhospital} & Iterative Evolution & - \\
    \cline{2-6}
     & \multicolumn{1}{|c|}{\multirow{3}{*}{Medical Training}} & \multicolumn{1}{|c|}{\multirow{2}{*}{Training Environment}} & \cite{wei2024medcomedicaleducationcopilots} & Collaborative Experts & Yes \\
    \cline{4-6} 
     & & \multicolumn{1}{|c|}{} & \cite{wu2024surgboxagentdrivenoperatingroom} & Collaborative Experts & - \\
    \cline{3-6} 
     & & \multicolumn{1}{|c|}{Scenario Simulation} & \cite{yu2024aipatientsimulatingpatientsehrs} & Collaborative Experts & Yes \\
    \cline{1-6}
    \multirow{3}{8em}{Healthcare Service Optimization} & \multicolumn{2}{|c|}{\multirow{2}{*}{Automation of Non-diagnostic Tasks}} & \cite{mukherjee2024polarissafetyfocusedllmconstellation} & Sequential Task Chain & Yes \\
    \cline{4-6}
     & && \cite{laymouna2024roles} & -- & - \\
    \cline{2-6}
     & \multicolumn{2}{|c|}{Automation of Diagnostic Tasks} & \cite{chadebecq2023artificial} & Iterative Evolution & - \\
\bottomrule
\end{tabular}
}
\label{tab:appliction}
\end{table*}

\subsection{Clinical Decision Support and Diagnosis}
In the area of Clinical Decision Support and Diagnosis, multi-agent frameworks based on LLMs improve clinical decision-making by addressing limitations of standalone LLMs. Systems in this area assign specialized roles to agents for intent recognition, diagnostic reasoning, and treatment planning so that healthcare delivery can be both personalized and sensitive to the context. For example, the framework proposed by Dutta and Hsiao \cite{dutta2024adaptivereasoningactingmedical} simulates interactions between doctors and patients to refine diagnostic reasoning and has shown better performance on datasets such as MedQA. The system developed by Ke et al. \cite{info:doi/10.2196/59439} reduces cognitive biases in diagnosis by using agents that provide expert opinions and critical evaluations. Other systems, such as \textit{MedAide} \cite{wei2024medaideomnimedicalaide}, coordinate agents across stages including pre-diagnosis, diagnosis, medication, and post-diagnosis, while frameworks such as MDagents \cite{kim2024demonstrationadaptivecollaborationlarge} and EHRagent \cite{tang2024medagentslargelanguagemodels} improve diagnostic accuracy through structured discussions and shared reasoning. Domain-specific applications also show promise. For instance, the work by Yue et al. \cite{yue2024clinicalagentclinicaltrialmultiagent} uses multi-agent collaboration to predict clinical trial outcomes by integrating large-scale domain knowledge. The Polaris framework \cite{mukherjee2024polarissafetyfocusedllmconstellation} combines general communication agents with task-specific agents to ensure safe patient interactions, and the system known as \textit{Rx Strategist} \cite{van2024rxstrategistprescriptionverification} uses knowledge graphs and multi-stage reasoning to check prescriptions for correct indications, dosages, and drug interactions.

\subsection{Clinical Data Analytics and Documentation}
In Clinical Data Analytics and Documentation, LLM-based agents show strong performance in processing both structured and unstructured data by using advanced architectures and retrieval-augmented generation techniques. The system \textit{ColaCare} proposed by Wang et al. \cite{wang2024colacareenhancingelectronichealth} integrates different agents to perform tasks such as mortality prediction and analysis of hospital readmission, demonstrating improved performance on the MIMIC-III and MIMIC-IV datasets. The work by Lee et al. \cite{lee2024improvingclinicaldocumentationai} introduces Sporo AI Scribe to address challenges related to the variability and complexity of clinical documentation. Research by Sudarshan \cite{sudarshan2024agenticllmworkflowsgenerating} shows that technical medical reports can be converted into patient-friendly formats by using iterative self-reflection and retrieval-augmented generation. In addition, \textit{Agent Hospital} \cite{li2024agenthospitalsimulacrumhospital} contributes to simulation systems by generating complete interactions that improve diagnostic and treatment capabilities.

\subsection{Medical Training and Simulation}
In Medical Training and Simulation, simulation environments are used to test and refine LLM-based agents before their use in clinical practice. Systems such as \textit{ClinicalLab} \cite{yan2024clinicallabaligningagentsmultidepartmental} and \textit{AI Hospital} \cite{fan2024aihospitalbenchmarkinglarge} evaluate diagnostic and treatment performance by simulating interactions across many specialties and complex healthcare scenarios. The system \textit{Agent Hospital} \cite{li2024agenthospitalsimulacrumhospital} further improves this process by allowing repeated training through large-scale simulations. In the field of medical education, systems such as \textit{MEDCO} \cite{wei2024medcomedicaleducationcopilots} support training in diagnostic reasoning and collaborative problem solving, while \textit{AIPatient} \cite{yu2024aipatientsimulatingpatientsehrs} integrates electronic health records with knowledge graphs to simulate realistic clinical scenarios. The system \textit{SurgBox} \cite{wu2024surgboxagentdrivenoperatingroom} provides a training environment for surgical procedures with real-time decision support that has been validated against actual surgical records.

\subsection{Healthcare Service Optimization}
In Healthcare Service Optimization, LLM-based agents improve the delivery of healthcare by automating tasks such as patient education, data collection, and support services. Research shows that automating these non-diagnostic tasks reduces the workload of healthcare professionals while maintaining service quality \cite{swarmsofLLM2024,mukherjee2024polarissafetyfocusedllmconstellation,laymouna2024roles}. There is also potential for the future automation of certain diagnostic tasks, including endoscopies and surgeries \cite{chadebecq2023artificial}. These implementations have resulted in measurable improvements in operational efficiency and patient satisfaction.

\section{Evaluation and Benchmarking} \label{sec:evaluation}

Evaluating LLM-based medical agents is essential for confirming their reliability, safety, and clinical effectiveness. A comprehensive evaluation framework is required to measure performance across different medical tasks, identify limitations, and guide improvements for clinical applications. A summary of the evaluation metrics and benchmark categories is provided in Table~\ref{tab:eval}.

\begin{table*}[htbp]
\renewcommand\arraystretch{1.2}
\centering
\caption{Common Evaluation Benchmarks and Metrics}
\scalebox{0.9}{
\begin{tabular}{c|c|l|l}
    \cline{1-4}
    \multirow{1}{*}{\textbf{Evaluation Attribute}} & \textbf{Genre} & \textbf{Specific Names} & \textbf{Related Work} \\
    \cline{1-4}
    \multirow{11}{*}{Benchmarks} & \multirow{5}{10em}{Static Q\&A Benchmarks} & \textit{MedQA} & \cite{jin2020diseasedoespatienthave} \\
    \cline{3-4}
     & & \textit{MedMCQA} & \cite{pmlr-v174-pal22a} \\
    \cline{3-4}
     & & \textit{Pub-MedQA} & \cite{jin2019pubmedqa} \\
    \cline{3-4}
     & & \multirow{2}{*}{\textit{MMLU}} & \cite{hendryckstest2021} \\
     & & & \cite{hendrycks2021ethics} \\
    \cline{2-4}
     & \multirow{4}{10em}{Workflow-Based Simulation Benchmarks} & \textit{MedChain} & \cite{liu2024medchainbridginggapllm} \\
    \cline{3-4}
     & & \textit{AI Hospital} & \cite{fan2024aihospitalbenchmarkinglarge} \\
    \cline{3-4}
     & & \textit{AgentClinic} & \cite{schmidgall2024agentclinic} \\
    \cline{3-4}
     & & \textit{ClinicalLab} & \cite{yan2024clinicallabaligningagentsmultidepartmental} \\
    \cline{2-4}
     & \multirow{2}{10em}{Automated Evaluation Frameworks} & \textit{AI-SCE} & \cite{Mehandru2024EvaluatingLL} \\
    \cline{3-4}
     & & \textit{RJUA-SPs} & \cite{10.1145/3637528.3671575} \\ 
    \cline{1-4}
    \multirow{9}{7em}{Metrics for Task-Specific Evaluation} & \multirow{3}{10em}{Exact Match Metrics} & Accuracy & -- \\
    \cline{3-4}
     & & Precision & -- \\
    \cline{3-4}
     & & Recall & -- \\
    \cline{2-4}
     & \multirow{3}{10em}{Semantic Similarity Metrics} & \textbf{BLEU} & \cite{papineni-etal-2002-bleu} \\
    \cline{3-4}
     & & \textbf{ROUGE} & \cite{lin-2004-rouge} \\
    \cline{3-4}
     & & \textbf{BertScore} & \cite{zhang2020bertscoreevaluatingtextgeneration} \\
    \cline{2-4}
     & \multirow{3}{10em}{LLM-Based Evaluation Metrics} & \textit{ChatCoach} & \cite{huang2024benchmarkinglargelanguagemodels} \\
    \cline{3-4}
     & & \multirow{2}{10em}{Retrival-Augmented Evaluation Framework} & \multirow{2}{*}{\cite{10.1145/3637528.3671575}} \\ 
    &&&\\
    \cline{1-4}
\end{tabular}
}
\label{tab:eval}
\end{table*}
\subsection{Benchmark Categories}

Benchmarks for LLM-based medical agents can be divided into three categories. 

\noindent\textbf{Static Question-Answering} benchmarks evaluate medical knowledge through tasks that have predetermined answers. For example, MedQA \cite{jin2020diseasedoespatienthave} simulates USMLE-style questions, MedMCQA \cite{pmlr-v174-pal22a} includes 194,000 questions covering 2,400 topics across 21 subjects, PubMedQA \cite{jin2019pubmedqa} assesses the understanding of biomedical research, and MMLU \cite{hendryckstest2021,hendrycks2021ethics} offers cross-domain single-choice questions. Although these datasets are useful for testing factual knowledge, they do not capture the interactive and sequential decision-making seen in clinical practice.

\noindent\textbf{Workflow-based Simulation} benchmarks mimic clinical decision-making through multiple stages. For instance, MedChain \cite{liu2024medchainbridginggapllm} contains 12,163 cases from 19 specialties and uses 7,338 medical images, AI Hospital \cite{fan2024aihospitalbenchmarkinglarge} evaluates interactions between healthcare providers and patients using the MVME dataset, AgentClinic \cite{schmidgall2024agentclinic} offers versions for both multimodal analysis and dialogue-based scenarios, and ClinicalLab \cite{yan2024clinicallabaligningagentsmultidepartmental} tests diagnostic performance across 24 departments and 150 diseases. These benchmarks reflect the dynamics of clinical reasoning and the adaptation required when patient information changes, although their complexity makes standardization challenging.

\noindent\textbf{Automated Evaluation} frameworks are developed to reduce reliance on human evaluators. For example, AI-SCE \cite{Mehandru2024EvaluatingLL} uses an OSCE-based framework for systematic evaluation, and RJUA-SPs \cite{10.1145/3637528.3671575} applies automated evaluation methods in urology using standardized patients and retrieval-augmented techniques.

\subsection{Metrics for Task-specific Evaluation}

\textbf{Exact Match Metrics} are used for tasks with clear correct answers, such as multiple-choice questions. In these tasks, accuracy, precision, and recall are calculated by directly comparing the model outputs with reference answers. Benchmarks such as MedQA \cite{jin2020diseasedoespatienthave} and MedMCQA \cite{pmlr-v174-pal22a} often use these metrics. While these metrics are effective for assessing factual knowledge, they may not be sufficient for tasks that involve complex reasoning or detailed explanations.

\noindent\textbf{Semantic Similarity Metrics} are applied to text generation tasks, such as writing clinical reports or diagnostic summaries. These metrics assess how well the meaning of the generated text matches that of the reference text. Metrics such as BLEU \cite{papineni-etal-2002-bleu}, which measures n-gram overlap, ROUGE \cite{lin-2004-rouge}, which evaluates summarization quality, and BertScore \cite{zhang2020bertscoreevaluatingtextgeneration}, which uses contextual embeddings to capture semantic relationships, have been applied in benchmarks such as ClinicalLab and MedChain.

\noindent\textbf{LLM-based Evaluation Metrics} use language models themselves to evaluate outputs based on factors such as coherence, relevance, and reasoning quality. For example, ChatCoach \cite{huang2024benchmarkinglargelanguagemodels} uses LLMs to assess the effectiveness of communication and decision making in patient consultations, while the Retrieval-Augmented Evaluation framework \cite{10.1145/3637528.3671575} used in RJUA-SPs measures the alignment of outputs with standard clinical pathways. This approach provides a scalable and adaptable method for assessing complex, multi-step clinical tasks.

\section{Discussions}
\label{sec:discussion}

Integrating Large Language Model (LLM)-based agents into medical workflows presents both challenges and opportunities. While previous work has achieved successes, the field remains in its early stages. Several significant challenges persist, and many opportunities require further exploration to fully realize their potential in healthcare applications. The following sections discuss these challenges and opportunities.

\subsection{Technical Challenges}

\subsubsection{Hallucination Management} LLM hallucinations—instances where models generate incorrect or misleading information—pose a significant risk in medical contexts, potentially leading to erroneous diagnoses and treatments \cite{Huang_2024}. Benchmarks such as \textit{MedHallBench} \cite{zuo2024medhallbench} and \textit{HaluEval} \cite{li2023halueval} highlight the need for reliable verification systems and error prevention mechanisms, especially in multi-agent scenarios where mistakes can propagate. Future research should focus on developing verification systems and dynamic error-correction methods that continuously update models with real-time, validated medical knowledge.

\subsubsection{Multimodal and Multilingual Integration} LLM-based agents must process various data types, including clinical texts and medical images, and handle variability in medical terminology across different languages and cultures \cite{li2024mmedagentlearningusemedical,Mehandru2024EvaluatingLL}. Variations in documentation standards and regional practices add to this complexity. It is crucial to develop models that can reliably operate in both multilingual and multimodal contexts.

\subsubsection{Cross-Department Integration} Healthcare environments encompass various departments, such as emergency, outpatient, and long-term care, each with its own workflows and documentation standards \cite{Qiu2024}. Achieving interoperability and accurate data exchange among these settings is challenging. Future work should focus on developing universal standards and adaptive interfaces that harmonize terminology and processes across departments, ensuring effective communication among LLM-based agents.

\subsection{Evaluation Challenges} Evaluating LLM-based medical agents is challenging. Traditional static benchmarks, which focus on fixed question-answering tasks, do not capture the dynamic and interactive aspects of clinical workflows, such as sequential decision-making, adaptive reasoning, and effective communication with patients and clinicians \cite{jin2020diseasedoespatienthave,schmidgall2024agentclinic}. Moreover, many medical applications require integrating heterogeneous data types, including text records, images, and laboratory results, which calls for evaluation frameworks that accurately simulate multimodal interactions \cite{liu2024medchainbridginggapllm,fan2024aihospitalbenchmarkinglarge}. Standard language metrics like BLEU \cite{papineni-etal-2002-bleu} and ROUGE \cite{lin-2004-rouge} assess only textual overlap and do not reflect clinical outcomes such as diagnostic accuracy. Additionally, dataset biases—such as the overrepresentation of specific conditions—can limit the generalizability of evaluation results across different healthcare settings \cite{yan2024clinicallabaligningagentsmultidepartmental}. Future research should develop integrated, multimodal evaluation frameworks that combine quantitative measures with qualitative clinical assessments and establish standardized clinical performance metrics while reducing dataset biases \cite{Mehandru2024EvaluatingLL}.

\subsection{Implementation Barriers}


\subsubsection{System Integration Complexity}
Large-scale systems like the Polaris healthcare system \cite{mukherjee2024polarissafetyfocusedllmconstellation}, which involve millions of professionals and established decision-making processes, illustrate the complexity of integration. Although many LLM frameworks prove valuable in specific applications, their broader integration does not always lead to improvements in operational efficiency.

\subsubsection{Resource Allocation Dilemma}
Developing and maintaining LLM-based agents requires significant computational resources, resulting in high costs for medical institutions. Such investments may produce systems that are not entirely reliable, raising concerns about their cost-effectiveness.

\subsection{Ethical and Privacy Concerns}
\subsubsection{Patient-Centered Design}
Medical diagnosis systems powered by LLM agents currently receive limited feedback from patients and caregivers, despite the importance of including their perspectives in decision-making \cite{kim2024mdagents}. Most existing frameworks focus solely on interactions with physicians. A more responsible approach would integrate patient narratives, physician observations, and caregiver input to support a truly patient-centered process.

\subsubsection{Algorithmic Bias}
Both general-purpose and medically fine-tuned LLMs can exhibit various biases, including social and cognitive biases. The \textit{BiasMedQA} benchmark \cite{schmidgall2024addressing} evaluated seven types of bias in state-of-the-art medical LLMs and found that precision can fall below 80\%, with some models performing as poorly as 50\%. This raises concerns about the reliability of these models in addressing bias. Medical agents must be designed to make responsible decisions, and reducing bias is essential for achieving this goal.

\subsubsection{Privacy and Security Threats}
Sensitive data used for training may be exposed during text generation or extracted through techniques such as inference attacks \cite{kandpal2023user} or data extraction \cite{carlini2021extracting}. It is critical to protect sensitive information in accordance with regulations like GDPR (EU) \cite{gdpr2016general} and HIPAA (USA) \cite{act1996health} when deploying medical agents. Data collection for developing LLM agents must prioritize privacy protection. \citep{dou2024exploringllmbaseddataannotation} suggests using LLMs for autonomous data generation and labeling as a means to protect privacy. In addition, privacy-preserving data processing methods, such as differential privacy, can add controlled noise to data so that individual records do not significantly influence overall results while preserving data utility.

\subsection{Future Opportunities and Application}

\subsubsection{Inspiration of O1 and R1 for Medical Reasoning}
The evolution of LLM-based medical agents can benefit from insights drawn from DeepSeek R1 and inference-time scaling strategies. DeepSeek R1 \cite{guo2025deepseek} has shown that reinforcement learning combined with long-chain reasoning leads to more accurate and context-aware medical decision-making, offering potentials for improving autonomous medical agents \cite{faray2025does}. By continuously optimizing AI-generated diagnoses and treatment recommendations through iterative self-evolution, LLM-based agents can better integrate multimodal clinical data, including electronic health records, medical images, and laboratory findings \cite{xu2024nextgenerationmedicalagento1}. Inference-time scaling, allowing LLMs more reasoning time, has been shown to improve performance in complex tasks such as differential diagnosis and treatment planning \cite{huang2025o1replicationjourney}, consistent with the hypothetico-deductive method used in clinical reasoning. Future research should explore how LLM-based agents can dynamically adjust inference time based on task complexity while incorporating reinforcement learning-based optimization techniques to enhance adaptability and reliability in clinical settings.

\subsubsection{Integration with Physical Systems}
Expanding LLM-based agents from virtual applications to integration with physical systems represents a significant step in medical care. While LLMs excel at data analysis and decision support, connecting them with physical systems such as medical robots could enable direct patient care. Such systems might combine language processing with physical inputs to support tasks like surgical assistance and patient monitoring. For example, empowering nursing robots \cite{nursingrobot} is one potential approach. However, this integration raises challenges regarding safety and real-time performance. Addressing technical limitations, ensuring system reliability, and resolving ethical concerns are necessary for successful integration. Hardware systems must accurately execute LLM outputs because errors could endanger patient safety, and high costs or technical complexity may limit system availability. Future work should focus on improving the integration of LLM-based agents with physical systems and on creating practical implementation frameworks.

\subsubsection{Advancements in Training Simulation}
Current medical LLM agents often use simulated hospitals for training, such as the \textit{Agent Hospital} framework, which enables the autonomous evolution of doctor agents through synthetic patient interactions \cite{li2024agenthospitalsimulacrumhospital}. Extending these simulations to include educational medical games could improve training data generation and learning experiences, even though challenges in data quality remain. AI-driven patient simulations that provide structured feedback have demonstrated effectiveness in enhancing clinical decision-making \cite{Brgge2024LargeLM}, but validating data generated by such games remains resource intensive.

\section{Conclusion}

This survey examines LLM-based agents in medicine, covering their architectures, applications, and challenges. While these agents enhance diagnostics, data analysis, and clinical workflows, issues remain in hallucination management, multimodal integration, and medical reasoning accuracy. Future work should focus on real-time error correction, improved multimodal fusion, and hybrid reasoning to enhance reliability and clinical utility.

\section*{Limitations}
This survey has several inherent limitations that should be considered. Due to the rapid development of LLM-based medical agents, our review primarily covers works published between 2022 and early 2024, which means future developments may introduce new architectures and approaches not captured in this analysis. Additionally, while we aimed for comprehensive coverage, we focused mainly on English-language publications in major academic databases such as PubMed, ACM Digital Library, arXiv, and Google Scholar. Valuable work published in other languages or regional databases may not be included in our analysis. These limitations reflect the inherent constraints of conducting a survey in a rapidly evolving field rather than shortcomings in the reviewed research itself.




\bibliography{custom}

\appendix

\section{Appendix}
\label{sec:appendix}

\subsection{Paper Collection}
All papers included in this review were identified through a systematic search of major academic databases such as PubMed, ACM Digital Library, arXiv, and Google Scholar. Keywords such as "large language model", "medical agent", "clinical decision support", and "healthcare AI" were used to select relevant studies published between 2022 and 2024. This process initially identified about 300 articles, from which 80 were shortlisted based on title and abstract screening for relevance and quality. After a full-text review, approximately 60 studies specifically addressing LLM-based agents in medical contexts were selected for this survey.

\subsection{Evaluation and Benchmarking}
Table~\ref{tab:eval} demonstrates common evaluation benchmarks and metrics for the llm-based agents in medicine.
\end{document}